\documentclass[runningheads]{llncs}

\usepackage[utf8]{inputenc}
\usepackage{graphicx}
\usepackage{url}
\usepackage{xcolor} 
\usepackage{hyperref}
\usepackage{xspace}
\usepackage{tabularx}
\usepackage{balance}

\usepackage{amsmath}
\usepackage{amssymb}
\DeclareMathOperator*{\eqdef}{\stackrel{\text{\tiny def}}{=}}
\DeclareMathOperator*{\argmax}{arg\,max}
\DeclareMathOperator*{\tf}{Tf}
\DeclareMathOperator*{\idf}{Idf}
\DeclareMathOperator*{\tfidf}{\tf\!\cdot\!\idf}
\newcommand{\R}{\mathbb{R}}

\graphicspath{{Figs/}}

\begin{document}
\begin{frontmatter}
\title{Open Set Classification of Untranscribed Handwritten Documents}

\author{  Jos\'e Ram\'on Prieto\inst{1}\orcidID{0000-0002-1207-6024} \and
  Juan Jos\'e Flores\inst{1}\orcidID{0000-0001-5385-5107} \and
  Enrique Vidal\inst{3}\orcidID{0000-0003-4579-5196} \and
  Alejandro H. Toselli\inst{1} \and
  David Garrido\inst{2}\orcidID{0000-0003-4617-1078} \and\\
  Carlos Alonso\inst{3}\orcidID{0000-0002-4130-3708}
  }
\authorrunning{J.\,R.\,Prieto et al.}

\institute{PRHLT Research Center,
  Universitat Polit\`ecnica de Val\`encia, Spain
\and HUM313 Research Group, Universidad de C\'adiz \and tranSkriptorium IA}

\maketitle
%
\begin{abstract}
  Huge amounts of digital page images of important manuscripts are
  preserved in archives worldwide.  The amounts are so large that it
  is generally unfeasible for archivists to adequately tag most of
  the documents with the required metadata so as to allow proper
  organization of the archives and effective exploration by scholars
  and the general public.
  The class or ``typology'' of a document is perhaps the most
  important tag to be included in the metadata. 
  The technical problem is one of automatic classification of
  documents, each consisting of a set of untranscribed handwritten
  text images, by the textual contents of the images.
  The approach considered is based on ``probabilistic indexing'', a
  relatively novel technology which allows to effectively represent
  the intrinsic word-level uncertainty exhibited by handwritten text
  images.
  We assess the performance of this approach on a large collection of
  complex notarial manuscripts from the %
  \textit{Spanish Archivo Host\'orico Provincial de C\'adiz},
  with promising results.
\end{abstract}


\end{frontmatter}

%
\section{Introduction}
Content-based classification of manuscripts is an important task that
is generally performed by expert archivists.  Unfortunately, however,
many manuscript collections are so vast that it is not possible to
have the huge number of archive experts that would be needed to
perform this task.

Current approaches for textual-content-based manuscript
classification require the handwritten images to be first
transcribed into text -- but achieving sufficiently accurate
transcripts are generally unfeasible for large sets of historical
manuscripts.  We propose a new approach to perform automatically
this classification task which does not rely on any explicit image
transcripts.

Hereafter, bundles or folders of manuscript images are called ``image
bundles'' or just ``bundles'' or ``books''.  A bundle may contain
several ``files'', also called ``acts'' or just ``image documents''.
The task consists of classifying a given image document, that may
range from a few to tens of handwritten text images, into a predefined
set of classes or ``types''.  Classes are associated with the topic or
(semantic) content conveyed by the text written in the images of the
document.

This task is different from other related tasks which, are often called
with similar names, such as ``content-based image classification'',
applied to single, natural scene (not text) images, and ``image
document classification'', where classification is based on visual
appearance or page layout.  See~\cite{prieto2021} for a more detailed
discussion on these differences, as well as references to previous
publications dealing with related problems, but mainly aimed at
printed text.

Our task is comparable to the time-honoured and well known task of
\emph{content-based document classification}, were the data are plain
text documents.  Popular examples of this traditional task, are
\emph{Twenty News Groups}, \emph{Reuters}, \emph{WebKB},
etc.~\cite{manning2008,khan2010,aggarwal2012}.
The task here considered (textual-content-based handwritten text image
document classification), is similar, except for a severe difference:
our data are sets of digital images of handwritten text rather than
flile of (electronic) plain text.
The currently accepted wisdom to approach our task would be to split
the process into two sequential stages. First, a handwritten text
recognition (HTR) system is used to transcribe the images into text
and, second, content-based document classification methods, such as
those referred to above, can be applied to the resulting text
documents.

This approach might work to some extent for simple manu\-scripts,
where HTR can provide over 90\% word recognition
accuracy~\cite{sanchez2019}.
But it is not an option for large historical collections, where the
best available HTR systems can only provide word recognition
accuracies as low as 40-60\%~\cite{sanchez2019,romero2019,vidal2020}.
This is the case of the collection which motivates this work, which
encompasses millions of handwritten notarial files from the Spanish
Archivo Hist\'orico Provincial de C\'adiz.  A small subset of these
manuscripts was considered in the Carabela project~\cite{vidal2020}
and the average word recognition accuracy achieved
was below 65\%~\cite{romero2019}, dropping to 46\% or less when
conditions are closer to real-world usage~\cite{vidal2020}.
Clearly, for these kinds of manuscript collections, the aforementioned
two-stage idea would not work and more holistic approaches are needed.

In previous works~\cite{vidal2020,prieto2021}, we have proposed an
approach which strongly relies on the so-called
\emph{probabilistic indexing} (PrIx) technology, recently developed
to deal with the intrinsic word-level \emph{uncertainty} generally
exhibited by handwritten text and, more so, by historical handwritten
text
images~\cite{toselli2016,toselli2017,lang2018,puigcerver2018,toselli2019}.
%
%
This technology was primarily developed to allow search and retrieval
of textual information in large untranscribed manuscript
collections~\cite{toselli2017,toselli2019a,vidal2020}.

In our proposal, PrIx provides the probability distribution of words
which are likely written in the images, from which statistical
expectations of \emph{word} and \emph{document} \emph{frequencies}
are estimated.  These estimates are then used to compute well-known
text features such as \emph{Information Gain} and
$\tfidf$~\cite{manning2008}, which are in turn considered inputs to a
Multilayer Perceptron classifier~\cite{prieto2021}.

In this paper, we further develop and consolidate this approach and,
as mentioned above, apply it to a new collection of handwritten
notarial documents from the Archivo Provincial de C\'adiz.
In contrast with~\cite{vidal2020,prieto2021}, where the underlying
class structure was very limited (just three rather artificial
classes), here the classes correspond to real typologies,
such as \emph{power of attorney}, \emph{lease}, \emph{will}, etc.
Our results clearly show the capabilities of the proposed approach,
which achieves classification accuracy as high as 90--97\%, depending
on the specific set of manuscripts considered.

\section{Probabilistic Indexing of Handwritten Text Images}
\vspace{-0.2em}
\label{sec:PrIx}

The Probabilistic Indexing (PrIx) framework was proposed to deal with
the intrinsic word-level 
uncertainty generally exhibited by handwritten text in images and, in
particular, images of historical manuscripts. It draws from ideas and
concepts previously developed for keyword spotting, both in speech
signals and text images.  However, rather than caring for "key" words,
any element in an image which is likely enough to be interpreted as a
word is detected and stored, along with its 
\emph{relevance probability} (RP) and its location in the image.
These text elements are referred to as \emph{``pseudo-word spots''}.


Following~\cite{toselli2016,puigcerver2018}, the image-region word RP
is denoted as $P(R\!=\!1\mid X\!=\!x, V\!=v\!)$, but for the sake of
conciseness, the random variable names will be omitted and, for $R=1$,
we will simply write $R$.
As discussed in~\cite{Vidal:17}, this RP can be simply approximated
as:
\vspace{0.3em}
\begin{equation}\label{eq:relevProb0}
P(R\mid x,v) = \sum_{b\sqsubseteq x} P(R,b\mid x,v)
       \approx \max_{b\sqsubseteq x} P(v\mid x,b)
\end{equation}
where $b$ is a small, word-sized image sub-region or Bounding Box
(BB), and with $b\sqsubseteq x$ we mean the set of all BBs contained
in $x$.
$P(v\mid x,b)$ is just the posterior probability needed to
``recognize'' the BB image $(x,b)$.
Therefore, assuming the computational complexity entailed
by~\eqref{eq:relevProb0} is algorithmically managed, any sufficiently
accurate isolated word classifier can be used to obtain
$P(R\mid x,v)$.

This word-level indexing approach has proved to be very robust, and it
has been used to very successfully index several large iconic
manuscript collections, such as the French \textsc{Chancery}
collection~\cite{toselli2017}, the \textsc{Bentham
  papers}~\cite{toselli2019a}, and the Spanish \textsc{Carabela}
collection considered in this paper, among others.\!%
\footnote{See: \url{http://transcriptorium.eu/demots/KWSdemos}}

\vspace{-0.3em}
\section{Plain Text Document Classification}\label{sec:docClass}
\vspace{-0.2em}

If a text document is given in some electronic form, its words can be
trivially identified as discrete, unique elements, and then the whole
field of \emph{text analytics}~\cite{manning2008,aggarwal2012} is
available to approach many document processing problems, including %
\emph{document classification} (DC).
%
%
Most DC methods assume a document representation model known
as~\emph{vector model} or~\emph{bag of words}
(BOW)~\cite{ikonomakis2005,manning2008,aggarwal2012}.  In this model,
the order of words in the text is ignored, and a document is
represented as a \emph{feature vector} (also called ``word
embedding'') indexed by $V$.
Let $\mathcal{D}$ be a set of documents, $D\in\mathcal{D}$ a document,
and $\vec{D}\in\R^N$ its BOW representation, where $N\eqdef|V|$.  For
each word $v\in V$, $D_v\in\R$ is the value of the $v$-th feature of
$\vec{D}$.

Each document is assumed to belong to a unique class $c$ out of a
finite number of classes, $C$.  The task is to predict the best class
for any given document, $D$.
%
Among many pattern recognition approaches suitable for this
task, from those studied in~\cite{prieto2021} the Multi-Layer
Perceptron (MLP) was the one most promising.

\vspace{0.3em}
\subsection{Feature Selection}\label{sec:IG}
\vspace{-0.2em}

Not all the words are equally helpful to predict the class of a
document $D$.  Thus, a classical first step in DC is to determine
a ``good'' vocabulary, $V_{n}$, of reasonable size $n<N$.
One of the best ways to determine $V_{n}$ is to compute the
\emph{information gain} (IG) of each word in $V$ and retain in $V_{n}$
only the $n$ words with highest IG.

Using the notation of~\cite{prieto2021}, let $t_v$ be the
value of a boolean random variable that is \emph{True} iff, for some
random $D$, the word $v$ appears in $D$. 
So, $P(t_v)$ is the probability that $\exists D\in\mathcal{D}$ such
that $v$ is used in $D$, and $P(\overline{t}_v) = 1-P(t_v)$ is the
probability that \emph{no} document uses $v$.
The IG of a word $v$ is then defined as: 
\vspace{-0.2em}
\begin{eqnarray}\label{eq:IG}
 \textrm{IG}(v) = 
   &-&\! \sum_{c\in{C}}{P(c)\log{P(c)}} \nonumber\\[+0.1em]
   &+& P(t_v) \sum_{c\in{C}}{P(c\mid t_v)\log{p(c\mid t_v)}} \nonumber\\[+0.1em]
   &+& P(\overline{t}_v) \sum_{c\in{C}}{P(c\mid\overline{t}_v)
                                       \log{P(c\mid\overline{t}_v)}}
\end{eqnarray}

\vspace{-0.1em}
\noindent
where $P(c)$ is de prior probability of class $c$, $P(c\mid t_v)$ is
the conditional probability that a document belongs to class $c$,
given that it contains the word $v$, and $P(c\mid\overline{t}_v)$ is
the conditional probability that a document belongs to class $c$,
given that it does \emph{not} contain $v$.
Note that the first addend of~Eq.\,\eqref{eq:IG} does not depend on
$v$ and can be ignored to rank all $v\in V$ in decreasing order of
IG$(v)$.

To estimate the relevant probabilities in Eq.\,\ref{eq:IG}, let
$f(t_v)\leq M\eqdef|\mathcal{D}|$ be the number of documents in
$\mathcal{D}$ which contain $v$ and $f(\overline{t}_v)=M-f(t_v)$ the
number of those which do \emph{not} contain $v$.
Let $M_c\leq M$ be the number of documents of class $c$, $f(c,t_v)$
the number of these documents which contain $v$ and
$f(c,\overline{t}_v)=M_c-f(c,t_v)$ the number of those that do
\emph{not} contain $v$.  Then, the relevant probabilities used
in~Eq.\,\eqref{eq:IG} can be estimated as follows:
%
\begin{align}\label{eq:IG_pt}
  ~~~~~~~~~~P(t_v) &~=~~~ \frac{f(t_v)}{M} ~~~~~~~~~~~~~~~~~~~~
  P(\overline{t}_v) &=~\frac{M-f(t_v)}{M}~~~~~~~~~~~\,\\[+0.4em]
  \!\!\!\!\!\!\!\!\!\!P(c\mid t_v) &~=~ \frac{f(c,t_v)}{f(t_v)}~~~~~~~~~~~~~~~~~
  P(c\mid \overline{t}_v)\!\!\!\!\!&=~  \frac{M_c-f(c,t_v)}{M-f(t_v)}~~~~~~~~
  \label{eq:IG_ptc}
\end{align}


%

\vspace{-1.2em}
\subsection{Feature Extraction}\label{sec:tfidf}
\vspace{-0.2em}

Using information gain, a vocabulary $V_{n}$ of size $n\leq N$ can be
defined by selecting the $n$ words with highest IG.  By attaching a
(real-valued) feature to each $v\in V_{n}$, a document $D$ can be
represente by a $n$-dimensional feature vector $\vec{D}\in\R^{n}\!$.

The value $D_v$ of each feature $v$ is typically related with the
frequency $f(v,D)$ of $v$ in $D$,
However, absolute word frequencies can dramatically vary with the size
of the documents and normalized frequencies are generally preferred.
Let $f(D)=\sum_{v\in V_{n}}\!\!f(v,D)$ be the total (or ``running'')
number of words in $D$.  The normalized frequency of $v\in V_{n}$,
often called \emph{term frequency} and denoted $\tf(v,D),$ is the
ratio $f(v,D)\,/\,f(D)$, which is is a max-likelihood estimate of the
conditional probability of word $v$, given a document $D$,
$P(v\!\mid\!D)$.

While $\tf$ adequately deals with document size variability, it has
been argued that better DC accuracy can be achieved by further
weighting each feature with a factor that reflects its ``importance''
to predict the class of a document.  Of course, IG could be used for
this purpose, but the so-called \emph{inverse document frequency}
($\idf$)~\cite{Salton1988,joachims1996,aizawa2003} is argued to be
preferable.  $\idf$ is defined as $\log(M\,/\,f(t_v))$, which,
according to Eq.\,\eqref{eq:IG_pt}, can be written as $-\log P(t_v)$.

Putting it all together, a document $D$ is represented by a feature
vector $\vec{D}$. The value of each feature, $D_v$, is computed as the
$\tfidf$ of $D$ and $v$; i.e., $\tf(v,D)$, weighted by $\idf(t)$:
%
\begin{align}\label{eq:tfidf_classic} \nonumber
~~~~~~~~~~~~~~~
D_v &= ~\tfidf(v,D)\!\! &=~ \tf(v,D) \cdot \idf(v)~~~~~~~~~~~~~~~~~~~~\\[+0.2em]
    &= ~P(v\!\mid\! D) \log\frac{1}{P(t_v)} \!\!
    &= ~\frac{f(v,D)}{f(D)} \log\frac{M}{f(t_v)}~~~~~~~~~~~~~~~~~~~
\end{align}

%
%

\vspace{-0.8em}
\section{Textual-Content-Based Classification of Sets of Images}
\label{sec:imageClassif}
\vspace{-0.2em}

The primary aim of PrIx is to allow fast and accurate search for
textual information in large image collections.  However, the
information provided by PrIx can be useful for many other text
analytics applications which need to rely on incomplete and/or
imprecise textual contents of the images.  In particular, PrIx results
can be used to estimate all the text features discussed in
Sec.\,\ref{sec:docClass}, which are needed for image document
classification.

\vspace{-0.3em}
\subsection{Estimating Text Features from Image PrIx's}
\vspace{-0.2em}
\label{sec:rw-estimation}

Since $R$ is a binary random variable, theRP $P(R\mid x,v)$ can be
properly seen as the statistical expectation that $v$ is written in
$x$.
As discussed in~\cite{prieto2021}, the sum of RPs for all the
pseudo-words indexed in an image region $x$ is the statistical
expectation of the number of words written in $x$.
Following this estimation principle, all the text features discussed
in Sec.\,\ref{sec:docClass}, which are needed for image document
classification can be easily estimated.

Let $n(x)$ be the total (or ``running'') number of words written in an
image region $x$ and and $n(X)$ the running words in an image document
$X$ encompassing several pages (i.e., $f(D)$, see \ref{sec:tfidf}).
Let $n(v,X)$ be the frequency of a specific (pseudo-)word $v$ in a
document $X$.  And let $m(v,\mathcal{X})$ be the number of documents
in a collection, $\mathcal{X}$, which contain the (pseudo-)word $v$.
As explained in~\cite{prieto2021}, the expected values of these counts
are:

\vspace{-1.2em}
\begin{eqnarray}\label{eq:numWords}
\hspace{4em}
E[n(x)]
 &=& 
                         \sum_{v} P(R\mid x,v)\!\!
\end{eqnarray}

\vspace{-0.5em}
\begin{equation}\label{eq:numWordsDoc}
\hspace{4em}
  E[n(X)]
  = \!\sum_{x\sqsubseteq X}\sum_{v} P(R\mid x,v)
\end{equation}

\vspace{-0.3em}
\begin{equation}\label{eq:wordFreqDoc}
\hspace{4em}
  E[n(v,X)] = \sum_{x\sqsubseteq X} P(R\mid x,v)
\end{equation}

\vspace{-0.3em}
\begin{equation}\label{eq:expectedDocsWithWord}
\hspace{4em}
  E[m(v,\mathcal{X})] = \sum_{X\sqsubseteq\mathcal{X}} \max_{x\in X} P(R\mid x,v) 
\vspace{-0.3em}
\end{equation}

\vspace{-0.3em}
\vspace{-0.2em}
\subsection{Estimating Information Gain and $\tfidf$ of Sets of Text Images}

Using the statistical expectations of document and word frequencies of
Eqs.\,(\ref{eq:numWords}--\ref{eq:expectedDocsWithWord}), IG and
$\tfidf$ can be strightforwardly estimated for a collection of text
images.
According to the notation used previously, a document $D$ in
Sec.\,\ref{sec:docClass} becomes a set of text images or
\emph{image document}, $X$.  Also, the set of all documents
$\mathcal{D}$ becomes the text image collection $\mathcal{X}$, and we
will denote $\mathcal{X}_c$ the subset of image documents of class
$c$.  Thus $M\eqdef|\mathcal{X}|$ is now the total number of image
documents and $M_c\eqdef|\mathcal{X}_c|$ the number of them which
belong to class $c$.

The document frequencies needed to compute the IG of a word, $v$ are
summarized in Eqs.\,(\ref{eq:IG_pt}--\ref{eq:IG_ptc}).
Now the number of image documents that contain the word $v$,
$f(t_v)\equiv m(v,\mathcal{X})$, is directly estimated using
Eq.\,\eqref{eq:expectedDocsWithWord},
and the number of image documents of class $c$ which contain
$v$, $f(c,t_v)$, is also estimated as in
Eq.\,\eqref{eq:expectedDocsWithWord} changing $\mathcal{X}$ with
$\mathcal{X}_c$.

On the other hand, the frequencies needed to compute the $\tfidf$
document vector features are summarized in
Eq.\,\eqref{eq:tfidf_classic}.  In addition to
$f(t_v)\!\equiv\!m(v,\mathcal{X})$, we need the total number of
running words in a document $D$, $f(D)$, and the number of times the
word $v$ appears in $D$, $f(v,D)$.
Clearly, $f(D)\!\equiv\! n(X)$ and $f(v,D)\!\equiv\! n(v,X)$, which
can be directly estimated using Eq.\,\eqref{eq:numWordsDoc}
and\,\eqref{eq:wordFreqDoc}, respectively.

\vspace{-0.3em}
\subsection{Image Document Classification}
\label{sec:imageDocClassif}\label{sec:MLPclassif}
\vspace{-0.2em}

Using the $\tfidf$ vector representation $\vec{X}$ of an image
document $X\in\mathcal{X}$, optimal prediction of the class of $X$ is
achieved under the minimum-error risk statistical framework as:
\vspace{-0.3em}
\begin{equation} \label{eq:BayesRule1}
\hspace{4em}
  c^\star(X) ~= \argmax_{c\in\{1,\dots,C\}}P(c\mid \vec{X})
\end{equation}

The posterior $P(c\mid \vec{X})$ can be computed following several
well-known approaches, some of which are discussed and tested
in~\cite{prieto2021}.  
Following the results reported in that paper, the Multi-Layer
Perceptron (MLP) was adopted for the present work.
The output of all the MLP architectures considered is a softmax layer
with $C$ units and training is performed by backpropagation using the
cross-entropy loss.  Under these conditions, it is straightforward
that the outputs for an input $\vec{X}$ approach $P(c\mid\vec{X})$,
$1\leq c\leq C$.  Thus Eq.\,\eqref{eq:BayesRule1} directly applies.

Three MLP configurations with different numbers of layers have been
considered.  In all the cases, every layer except the last one is
followed by batch normalization and ReLU activation
functions~\cite{Ioffe2015}.
The basic configuration is a plain $C$-class perceptron where the
input is totally connected to each of the $C$ neurons of the output
layer (hence no hidden layers are used).  For the sake of simplifying
the terminology, here we consider such a model as a ``0-hidden-layers
MLP'' and refer to it as MLP-0.
The next configuration, MLP-1, is a proper MLP including one hidden
layer with $128$ neurons.  The hidden layer was expected to do some
kind of intra-document clustering, hopefully improving the
classification ability of the last layer.
Finally, we have also tested a deeper model, MLP-2, with two hidden
layers with $128$ neurons each.

%
\section{Dataset and Experimental Settings} \label{sec:dataandexpsettings}
The dataset considered in this work is a small part of a huge
manuscript collection of manuscripos held by the Spanish Archivo Hist\'orico Provincial de
C\'adiz (AHPC).  In this section, we provide details of the dataset and
of the settings adopted for the experiments discussed in Sec.\ref{sec:expAndResults}.

\subsection{A Handwritten Notarial Document Dataset}
The AHPC (Provincial Historical Archive of C\'adiz) was established
in 1931, with the main purpose of collecting and guarding notarial
documentation that was more than one hundred years old. Its functions
and objectives include the preservation of provincial documentary
heritage and to offer a service to researchers that allows the use and
consultation of these documentary sources.

The notarial manuscripts considered in the present work come from a
very large collection of 16\,849 bundles or ``protocol books'', with
an average of 250 notarial acts or files and about 800 pages per book.

Among these AHPC books, 50 were included in the collection compiled in
the Carabela project~\cite{vidal2020}.%
\footnote{In \url{http://prhlt-carabela.prhlt.upv.es/carabela} the
  images of this collection and a PrIx-based search interface are
  available.}
From these 50 books, for the present work we selected two protocol
books, JMBD\_4949 and JMBD\_4950, dated 1723-1724, to be manually
tagged with GT annotations.

\begin{figure}[htpb]
\centering
\vspace{-2em}
 \includegraphics[width=\textwidth]{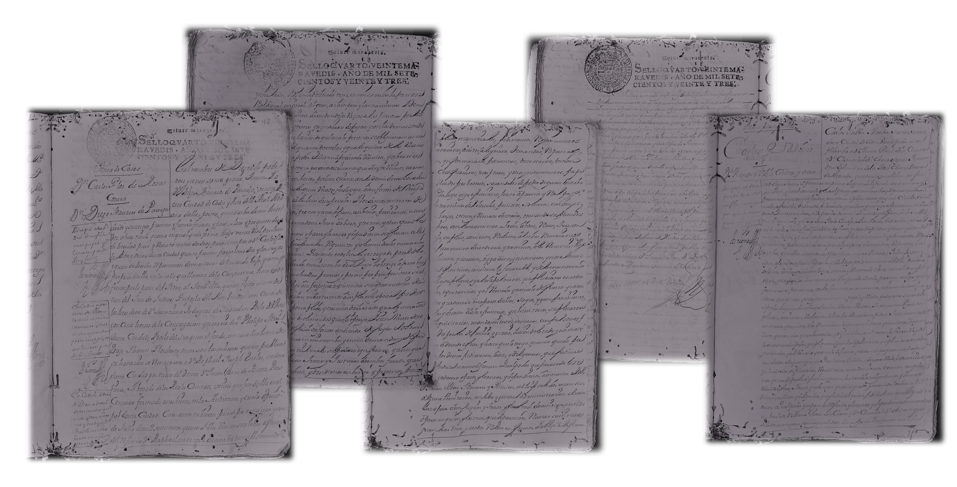}
 \vspace{-3em}
  \caption{Example of corpus pages from books JMDB\_4949 and JMBD\_4950.}
 \label{fig:ExPages}
\vspace{-1.5em}
\end{figure}

Figure \ref{fig:ExPages} shows examples of page images of these two
books.
The selected books were manually divided into sequential sections,
each corresponding to a notarial act.

A first section of about 50 pages, which contains a kind of table of
contents of the book, was also identified but not used in the present experiments.
It is worth noting that each notarial act can contain from one to
dozens of pages, and separating these acts is not straightforward.  In
future works, we plan to develop methods to also perform this task
automatically, but for the present work we take the manual
segmentation as given.

During the segmentation and labeling of the two protocol books, the
experts found a total of 558 notarial acts, 296 from JMBD\_4949 and 261 from JMBD\_4950, belonging to 38 different
types or classes.  However, for most classes, only very few acts were
available. To allow the classification results to be sufficiently
reliable, only those classes having at least \emph{five} acts in each
book were taken into account.

This way, \emph{thirteen classes} were retained as sufficiently representative and an 'other' class was added to represent the remaining classes, making it possible to use 557 notarial acts (i.e., documents), and discarding just one.(EXPLICAR PORQUE)

The thirteen types (\emph{classes}) we are finally left with are:
%
\emph{Power of Attorney} (P, from Spanish ``Poder''),
\emph{Letter of Payment} (CP, ``Carta de Pago''),
\emph{Debenture} (O, ``Obligaci\'on''),
\emph{Lease} (A, ``Arrendamiento''),
\emph{Will} (T, ``Testamento''),
\emph{Sale} (V, ``Venta''),
\emph{Risk} (R, ``Riesgo''),
\emph{Census} (CEN, ``Censo''),
\emph{Deposit} (DP, ``Deposito''),
\emph{Statement} (D, ``Declaraci\'on''),
\emph{Cession} (C, ``Cesi\'on''),
\emph{Treaty of fact} (TH, ``Tratado de hecho'') and
\emph{Redemption} (RED, ``Redenci\'on''),

Details of this dataset are shown in Table \ref{tab:data}.

\begin{table}[h!]
    \centering
    \vspace{-0.6em}
    \resizebox{\textwidth}{!}{%
    \extrarowheight=1pt\tabcolsep=3pt
    \begin{tabular}{c||ccccc}
          & \multicolumn{5}{c}{\large JMBD\_4949 \& JMBD\_4950} \\
    \hline
    \textbf{} \normalsize Classes & Nº docs. & Avg. pages per doc. & Min-max pages per doc. & Typical deviation & Total pages   \\
    \hline
    \normalsize
    P     & \normalsize 240    & \normalsize 3.34    & \normalsize 2-24    & \normalsize 3.45    & \normalsize 803   \\
    CP    & \normalsize 73     & \normalsize 4.78    & \normalsize 2-30    & \normalsize 5.38    & \normalsize 349   \\
    O     & \normalsize 44     & \normalsize 4.81    & \normalsize 2-32    & \normalsize 5.64    & \normalsize 212   \\
    A     & \normalsize 32     & \normalsize 4.75    & \normalsize 2-16    & \normalsize 2.63    & \normalsize 152   \\
    T     & \normalsize 29     & \normalsize 8.55    & \normalsize 4-48    & \normalsize 9.36    & \normalsize 248   \\
    V     & \normalsize 21     & \normalsize 22.85   & \normalsize 4-122   & \normalsize 29.79   & \normalsize 480   \\
    R     & \normalsize 17     & \normalsize 4.00    & \normalsize 4-4     & \normalsize 0.00    & \normalsize 68    \\
    CEN   & \normalsize 12     & \normalsize 11.50   & \normalsize 2-26    & \normalsize 9.02    & \normalsize 138   \\
    DP    & \normalsize 10     & \normalsize 3.80    & \normalsize 2-8     & \normalsize 1.88    & \normalsize 38    \\
    D     & \normalsize 10     & \normalsize 2.40    & \normalsize 2-4     & \normalsize 0.80    & \normalsize 24    \\
    C     & \normalsize 6      & \normalsize 5.33    & \normalsize 2-14    & \normalsize 3.94    & \normalsize 32    \\
    TH    & \normalsize 6      & \normalsize 5.33    & \normalsize 4-8     & \normalsize 1.88    & \normalsize 32    \\
    RED   & \normalsize 1      & \normalsize 12.00   & \normalsize 12-12   & \normalsize 0.0     & \normalsize 12    \\
    OTHER & \normalsize 56     & \normalsize 9.17    & \normalsize 2-70    & \normalsize 12.32   & \normalsize 514   \\
    \hline
    
    \textbf{\normalsize Total} & \textbf{\normalsize 557}   & \textbf{\normalsize 5.56}             & \textbf{\normalsize 2-122}         & \textbf{\normalsize 9.14}     & \textbf{\normalsize 3102}
    \end{tabular}
    }
\caption{Number of documents and page images for JMBD\_4949 and
    JMBD\_4950: per class, per document \& class, and totals.}
\vspace{-2em}
\label{tab:data}
\end{table}

The machine learning task consists in training a model to classify
each document into one of the $C=14$ classes considered.

\newpage

\subsection{Empirical settings} \label{sec:expSettings}
PrIx vocabularies typically contain huge amounts of pseudo-word
hypotheses.  However, many of these hypotheses have low relevance
probability and most of the low-probability pseudo-words are not real
words.
Therefore, as a first step, the huge PrIx vocabulary was pruned out
avoiding entries with less than three characters, as well as
pseudo-words $v$ with too low estimated document frequency; namely,
$E[m(v,\mathcal{X})]<1.0$.  This resulted in a vocabulary $V$ of
$559\,012$ pseudo-words for the two books considered in this work.
%
%
Secondly, to retain the most relevant features, (pseudo-)words were sorted by decreasing values
of IG and the first $n$ entries of the sorted list were selected to
define a BOW vocabulary $V_n$.  Exponentially increasing values of $n$
from $8$ up to $16\,384$ were considered in the experiments.
Finally, a $\tfidf$ $n$-dimensional vector was calculated for each
document, $D\equiv X\in \mathcal{X}$. For experimental simplicity,
$\tfidf(v,D)$ was estimated just once all for all $v\in V$, using the
normalized factor $f(D)\equiv E[n(X)]$ computed for all $v\in V$,
rather than just $v\in V_n$.

For MLP classification, document vectors were normalized by subtracting
the mean and dividing by the standard deviation, resulting in
zero-mean and unit-variance input vectors.
The parameters of each MLP architecture were initialized following
\cite{Glorot2010} and trained according to the cross-entropy loss for
100 epochs using the SGD optimizer \cite{Sebastian.R} with a learning
rate of $0.01$. This configuration has been used for all the
experiments presented in section \ref{sec:expAndResults}.
%
\section{Experiments and Results} \label{sec:expAndResults}
The empirical work has focused on MLP classification of documents
(handwritten notarial \emph{acts}) of two books, JMBD\_4949 and JMBD\_4950. For the two books at the same time, we classify their documents
(groups of images of handwritten pages) in the fourteen established classes in Sec.\,\ref{sec:dataandexpsettings}.

For this paper, two types of experiments have been performed. First, we performed a leave-one-out classification with the pages grouped by notarial acts and then, we performed another leave-one-out classification, but this time with the samples divided by pages and a subsequent vote to determine the class to which the notarial act belongs.
\newpage
\subsection{Groups} \label{sec:expgroups}
For this first experiment, we have made a leave one out classification with the pages grouped by notarial acts. 

\begin{figure}[htpb]
\centering
\vspace{-1.5em}
 \includegraphics[width=0.70\textwidth]{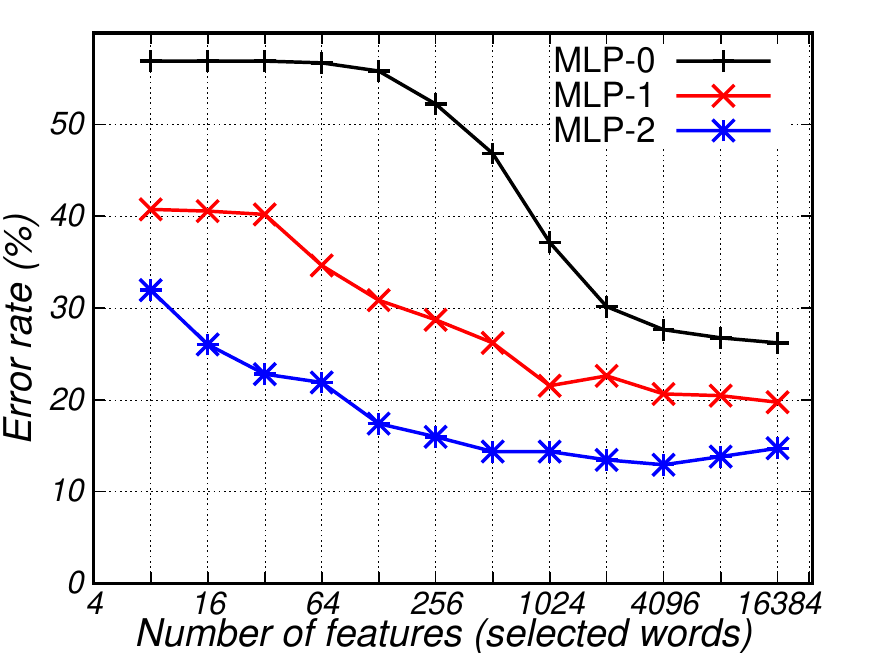}
 \vspace{-0.5em}
  \caption{Leaving-one-out classification error rate for three classifiers for JMBD\_4949 and JMBD\_4950.
    95\% confidence intervals (not shown for clarity) are all smaller
    than $\pm3.0\%$ for all the errors.}
 \label{fig:ExGroups}
\end{figure}

As shown in Figure \ref{fig:ExGroups}, best results are obtained using MLP-2, archieving an error of 12.92\% with a vocabulary of n = 4096 words. For this model we can see a degradation for lower and higher values of n. The rest of models not even come close to MLP-2 model, since the best result comes from MLP-1 and is 19.74 for n = 16384,  which is far from the best result obtained with MLP-2.

Model complexity, in terms of numbers of parameters to train, grows
with the number of features, $n$ as:

\begin{equation} \label{eq:numParam} \nonumber
\centering
\vspace{-0.2em}
  \text{MLP-0:}~ 5\,n + 5 ~~~~
  \text{MLP-1:}~ 128\,n + 773 ~~~~
  \text{MLP-2:}~ 128\,n + 17\,285 
  \vspace{-0.2em}
\end{equation}

For all $n>64$, the least complex model is MLP-0, followed by MLP-1
and \mbox{MLP-2}.  For $n=2\,048$, MLP-0, MLP-1 and MLP-2 have $10\,245$,
$262\,917$ and $279\,429$ parameters, respectively.
Therefore, despite the complexity of the model, MLP-2 is the best choice for the task considered in this work.

\newpage
Table ~\ref{tab:Confmat_groups} shows the average confusion matrix and the specific error rate per class, using the best model (MLP-2) with the best vocabulary ($n$).

\begin{table}[h]
\label{tab:Confmat_groups}
\begin{tabular}{c||rccccccccccccc|r|r}
\multicolumn{17}{c}{JMBD\_4949 \& JMBD\_4950}                                        \\
\hline
\textbf{} & \multicolumn{1}{c}{TH} & C & D & DP & CE & R & V & T & A & O & CP & P & RE & OT & \multicolumn{1}{c}{Total} & \multicolumn{1}{|c}{Error (\%)} \\
\hline
TH    & 6 & 0 & 0 & 0 & 0 & 0  & 0  & 0  & 0  & 0  & 0  & 0   & 0 & 0  & 6   & 0.0   \\
C     & 0 & 1 & 0 & 0 & 0 & 0  & 0  & 0  & 0  & 0  & 4  & 1   & 0 & 0  & 6   & 83.3  \\
D     & 0 & 0 & 3 & 0 & 0 & 0  & 0  & 3  & 1  & 0  & 0  & 0   & 0 & 3  & 10  & 70.0  \\
DP    & 0 & 0 & 0 & 9 & 0 & 0  & 1  & 0  & 0  & 0  & 0  & 0   & 0 & 0  & 10  & 10.0  \\
CEN   & 0 & 0 & 0 & 1 & 7 & 0  & 0  & 0  & 0  & 0  & 0  & 0   & 0 & 4  & 12  & 41.6  \\
R     & 0 & 0 & 0 & 0 & 0 & 17 & 0  & 0  & 0  & 0  & 0  & 0   & 0 & 0  & 17  & 0.0   \\
V     & 1 & 0 & 0 & 0 & 0 & 0  & 18 & 0  & 0  & 0  & 1  & 0   & 0 & 1  & 21  & 14.3  \\
T     & 0 & 0 & 0 & 0 & 1 & 0  & 0  & 26 & 0  & 0  & 0  & 1   & 0 & 1  & 29  & 10.3  \\
A     & 0 & 0 & 0 & 0 & 0 & 0  & 0  & 0  & 29 & 0  & 1  & 0   & 0 & 2  & 32  & 9.4   \\
O     & 0 & 0 & 0 & 0 & 0 & 0  & 0  & 0  & 1  & 33 & 1  & 3   & 0 & 6  & 44  & 25.0  \\
CP    & 0 & 0 & 0 & 0 & 0 & 0  & 2  & 0  & 1  & 3  & 62 & 3   & 0 & 2  & 73  & 15.06 \\
P     & 0 & 0 & 0 & 0 & 2 & 0  & 0  & 0  & 0  & 0  & 3  & 230 & 0 & 5  & 240 & 4.16  \\
RED   & 0 & 0 & 0 & 0 & 1 & 0  & 0  & 0  & 0  & 0  & 0  & 0   & 0 & 0  & 1   & 100.0 \\
OTHER & 0 & 0 & 1 & 0 & 4 & 0  & 2  & 0  & 0  & 3  & 3  & 4   & 0 & 39 & 56  & 30.3 
\end{tabular}

\end{table}

\newpage

\subsection{1-Page without voting} \label{sec:exppagenovote}

For this experiment, all the pages of the two files were classified one by one by means of leave one out.

\begin{figure}[htpb]
\centering
\vspace{-1.5em}
 \includegraphics[width=0.70\textwidth]{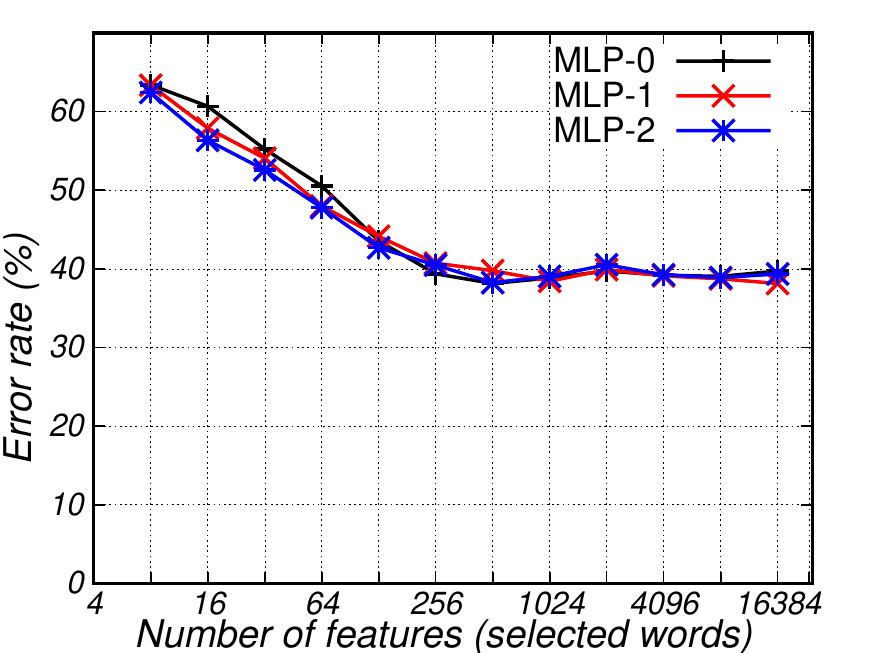}
 \vspace{-0.5em}
  \caption{Leaving-one-out classification error rate for three classifiers for JMBD\_4949 and JMBD\_4950.
    95\% confidence intervals (not shown for clarity) are all smaller
    than $\pm3.0\%$ for all the errors.}
 \label{fig:ExPagenovote}
\end{figure}

In this case, we can see in the Figure \ref{fig:ExPagenovote} a much higher error than in the previous experiment, since the pages have been sorted one by one and not by notarial acts. 
The best error rate is obtained with MLP-0 and MLP-1 for 512 and 16384 input words respectively, obtaining an error of 38.13\%.

From these results we can conclude that page-by-page classification models would not be of great advance to the field of content based document classification. However, if we add to these models a vote between the pages of the same notarial act to determine the class to which it belongs, things change, as we will see in the next section.

\newpage

\subsection{1-Page with voting} \label{sec:exppagevote}

Regarding this experiment, all the pages of the two files were classified one by one by means of leave one out, like the previous experiment, and then a vote was taken among the pages of each notarial act to determine to which class this act belongs.

\begin{figure}[h]
\centering
\vspace{-1.5em}
 \includegraphics[width=0.70\textwidth]{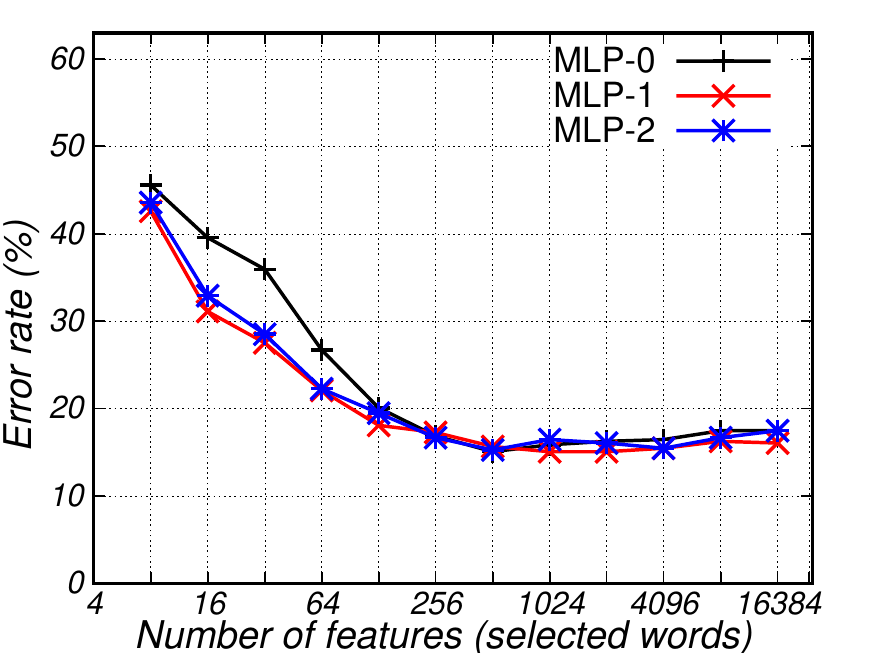}
 \vspace{-0.5em}
  \caption{Leaving-one-out classification error rate for three classifiers for JMBD\_4949 and JMBD\_4950 with subsequent voting.
    95\% confidence intervals (not shown for clarity) are all smaller
    than $\pm3.0\%$ for all the errors.}
 \label{fig:ExPage}
\end{figure}

As we can see in Figure \ref{fig:ExPage}, the results this time are much better than in the previous case. Here again we find a tie between MLP-0 and MLP-1, both obtaining the same error rate, although for different number of input words. 

While MLP-0 obtains an error rate of 15.06\% for n = 512, MLP-1 obtains this same rate for n = 1024 and n = 2048, which is why despite obtaining the same result, the MLP-1 model would be a better option for this work, since we can also verify that MLP-1 always obtains a lower error than MLP-0, except on two occasions (n = 256 and n = 512).

As a result of these error rates, we can conclude that voting significantly improves model predictions. Since we have gone from unproductive models to ones that would be of great utility and advance for expert archivists. 
%
\section{Conclusions} \label{sec:conclusions}

We have presented and showcased an approach that is able to perform
textual-content-based document classification directly on multi-page
documents of untranscribed handwritten text images.
Our method uses rather traditional techniques for plaintext document
classification, estimating the required word frequencies from image
probabilistic indexes.  This way, we overcome the need to explicitly
transcribe manuscripts, which is generally unfeasible for large
collections.

The experimental results obtained with the proposed approach leave no
doubt regarding its capabilities to model the textual contents of the
page images and to discriminate among content-defined classes.

In our opinion, probabilistic indexing opens new avenues for research
in textual-content-based image document classification.  
In future works, we plan to explore the use of other classification
methods based on information extracted from probabilistic indexes.
On the other hand, we aim to capitalize on the observation that fairly
accurate classification can be achieved with relatively small
vocabularies, down to 64 words in the task considered in this paper.
In this direction, we will explore the use of information gain and/or
$\tfidf$ values estimated for probabilistic index (pseudo-)words to
derive a small set of words that semantically describes the contents
of each bundle of manuscripts.  This would allow the automatic or
semi-automatic creation of metadata which could be extremely useful
for scholars and the general public searching for historical
information in archived manuscripts.

%
\balance

\section*{Acknowledgments}
This work has been supported by \dots

\bibliographystyle{splncs04}
\bibliography{docClassPaper}

\begin{thebibliography}{10}
\providecommand{\url}[1]{\texttt{#1}}
\providecommand{\urlprefix}{URL }
\providecommand{\doi}[1]{https://doi.org/#1}

\bibitem{aggarwal2012}
Aggarwal, C.C., Zhai, C.: Mining text data. Springer Science \& Business Media
  (2012)

\bibitem{aizawa2003}
Aizawa, A.: An information-theoretic perspective of tf--idf measures. Inf.
  Proc. {\&} Management  \textbf{39}(1),  45--65 (2003)

\bibitem{toselli2017}
Bluche, T., Hamel, S., Kermorvant, C., Puigcerver, J., Stutzmann, D., Toselli,
  A.H., Vidal, E.: {Preparatory KWS Experiments for Large-Scale Indexing of a
  Vast Medieval Manuscript Collection in the HIMANIS Project}. In: 14th IAPR
  Int. Conf. on Document Analysis and Recognition (ICDAR). vol.~01, pp.
  311--316 (Nov 2017)

\bibitem{vidal2020}
{E. Vidal et al.}: The carabela project and manuscript collection: Large-scale
  probabilistic indexing and content-based classification. In: 16th ICFHR (Sep
  2020)

\bibitem{Glorot2010}
Glorot, X., Bengio, Y.: {Understanding the difficulty of training deep
  feedforward neural networks}. Journal of Machine Learning Research
  \textbf{9},  249--256 (2010)

\bibitem{ikonomakis2005}
Ikonomakis, M., Kotsiantis, S., Tampakas, V.: Text classification using machine
  learning techniques. WSEAS transactions on computers  \textbf{4,8},  966--974
  (2005)

\bibitem{Ioffe2015}
Ioffe, S., Szegedy, C.: {Batch Normalization: Accelerating Deep Network
  Training by Reducing Internal Covariate Shift}  (2015)

\bibitem{joachims1996}
Joachims, T.: A probabilistic analysis of the {Rocchio} algorithm with {TFIDF}
  for text categorization. Tech. rep., Carnegie-mellon univ pittsburgh pa dept
  of computer science (1996)

\bibitem{khan2010}
Khan, A., Baharudin, B., Lee, L.H., Khan, K.: A review of machine learning
  algorithms for text-documents classification. Journal of advances in
  information technology  \textbf{1}(1),  4--20 (2010)

\bibitem{lang2018}
{Lang}, E., {Puigcerver}, J., {Toselli}, A.H., {Vidal}, E.: Probabilistic
  indexing and search for information extraction on handwritten german parish
  records. In: 2018 16th International Conference on Frontiers in Handwriting
  Recognition (ICFHR). pp. 44--49 (Aug 2018)

\bibitem{manning2008}
Manning, C.D., Raghavan, P., Schtze, H.: {Introduction to Information
  Retrieval}. Cambridge University Press, New York, NY, USA (2008)

\bibitem{prieto2021}
Prieto, J.R., Bosch, V., Vidal, E., Alonso, C., Orcero, M.C., Marquez, L.:
  Textual-content-based classification of bundles of untranscribed manuscript
  images. In: 2020 25th International Conference on Pattern Recognition (ICPR).
  pp. 3162--3169. IEEE (2021)

\bibitem{puigcerver2018}
Puigcerver, J.: A Probabilistic Formulation of Keyword Spotting. Ph.D. thesis,
  Univ. Politècnica de València (2018)

\bibitem{romero2019}
Romero, V., Toselli, A.H., Vidal, E., S{\'a}nchez, J.A., Alonso, C.,
  Marqu{\'e}s, L.: Modern vs diplomatic transcripts for historical handwritten
  text recognition. In: International Conference on Image Analysis and
  Processing. pp. 103--114. Springer (2019)

\bibitem{Sebastian.R}
Ruder, S.: An overview of gradient descent optimization algorithms
  \textbf{14}, ~2--3 (2017)

\bibitem{Salton1988}
Salton, G., Buckley, C.: {Term-weighting approaches in automatic text
  retrieval}. Inf. Proc. {\&} Management  \textbf{24}(5),  513/523 (1988)

\bibitem{sanchez2019}
S{\'a}nchez, J.A., Romero, V., Toselli, A.H., Villegas, M., Vidal, E.: A set of
  benchmarks for handwritten text recognition on historical documents. Pattern
  Recognition  \textbf{94},  122--134 (2019)

\bibitem{toselli2019a}
Toselli, A., Romero, V., Vidal, E., S\'anchez, J.: Making two vast historical
  manuscript collections searchable and extracting meaningful textual features
  through large-scale probabilistic indexing. In: 2019 15th IAPR Int. Conf. on
  Document Analysis and Recognition (ICDAR) (2019)

\bibitem{toselli2019}
Toselli, A.H., Vidal, E., Puigcerver, J., Noya-Garc{\'\i}a, E.: Probabilistic
  multi-word spotting in handwritten text images. Pattern Analysis and
  Applications  \textbf{22}(1),  23--32 (2019)

\bibitem{toselli2016}
Toselli, A.H., Vidal, E., Romero, V., Frinken, V.: {HMM Word Graph based
  Keyword Spotting in Handwritten Document Images}. Information Sciences
  \textbf{370-371},  497--518 (2016)

\bibitem{Vidal:17}
Vidal, E., Toselli, A.H., Puigcerver, J.: A probabilistic framework for
  lexicon-based keyword spotting in handwritten text images. Tech. rep.,
  {{UPV}} (2017)

\end{thebibliography}

\end{document}